\documentclass[smallextended]{svjour3} 
\usepackage{mathbbol}
\usepackage[numbers,sort&compress]{natbib}
\usepackage{graphicx}
\usepackage[colorlinks = true,
            linkcolor = blue,
            urlcolor  = blue,
            citecolor = blue,
            anchorcolor = blue]{hyperref}
\usepackage{textcomp}
\usepackage{subfigure}

\begin{document}
\title{Investigating the Parameter Space of Evolutionary Algorithms}

\author{Moshe Sipper \and 
Weixuan Fu 	 \and 
Karuna Ahuja \and 
Jason H. Moore}
				
\institute{Institute for Biomedical Informatics (IBI), 
Perelman School of Medicine,
University of Pennsylvania, Philadelphia, PA 19104. 
\email{\{sipper,weixuanf,karunaa,jhmoore\}@upenn.edu} 
\and
M. Sipper is also with the
Department of Computer Science, Ben-Gurion University, Beer-Sheva 8410501, Israel.
}

\date{Received: \today / Accepted: date}

\maketitle

\begin{abstract}
The practice of evolutionary algorithms involves the tuning of many parameters. 
How big should the population be? How many generations should the algorithm run? What is the (tournament selection) tournament size? What probabilities should one assign to crossover and mutation? Through an \textit{extensive} series of experiments over multiple evolutionary algorithm implementations and problems we show that parameter space tends to be rife with viable parameters, at least for 25 the problems studied herein. We discuss the implications of this finding in practice. 
\end{abstract}

\section{Introduction}
One of the crucial tasks of the evolutionary computation (EC) practitioner is the tuning of parameters. The fitness-select-vary paradigm comes with a plethora of parameters relating to the population, the generations, and the operators of selection, crossover, and mutation. It seems natural to ask whether the myriad parameters can be obtained through some clever methodology (perhaps even an evolutionary one) rather than by trial and error; indeed, as we shall see below, such methods have been devised. Our own interest in the issue of parameters stems partly from a desire to better understand evolutionary algorithms (EAs) and partly from our recent investigation into the design and implementation of an accessible artificial intelligence system \cite{Olson2017PennAI}.
 
In this paper we examine key parameters, asking whether we might improve the parameter-seeking process. EC practitioners often employ commonly used parameters ``selected by conventions, ad hoc choices, and very limited experimental comparisons'' \cite{eiben2011parameter}. We experimented with a large and variegated assortment of problems in what is arguably one of the most extensive EC experiments, concluding that parameter space, in fact, tends to be rife with viable parameters, at least for the 25 problems studied herein.  

We begin in the next section by delineating previous work on finding parameters through intelligent means. In Section~\ref{sec-GPs} we turn to our own experiments by first describing the software and datasets we used. Section~\ref{sec-megaton} presents our first approach for obtaining ``good'' parameter sets, which was based on a meta-genetic algorithm, followed by a random-search approach in Section~\ref{sec-rascal}. We discuss our findings and conclude in Section~\ref{sec-conc}.

\section{Previous Work: Seeking Parameters}
\label{sec-prev}

Before delving into a detailed discussion, let us make the distinction between \textit{parameters}, which are part of the model being evolved, and \textit{hyper-parameters} (also called \textit{meta-parameters}), which are not part of the model and need to be set by the user before running the evolutionary process, either manually or algorithmically. 
Examples of parameters are synaptic weights in deep neural networks and ephemeral random constants (ERCs) in genetic programming (GP), while examples of hyper-parameters include the number of hidden layers in deep neural networks and several standard parameters in GP: population size, generation count, crossover rate, mutation probability, and selection type.

\cite{brest2006self} noted that there are two major forms of setting \linebreak (hyper-)parameter values: \textit{parameter tuning} and \textit{parameter control}. Tuning refers to the common practice of seeking good values for the parameters before running the algorithm, then running the algorithm using these values, which remain fixed during the run. Parameter control means that values for the parameters are changed during the run.  

An early work by \cite{shahookar1990genetic} looked into the problem of VLSI layout, aiming to minimize the overall connection length between the components (cells) of a circuit. They devised a genetic algorithm that used three operators: crossover (order, cycle, or partially mapped---PMX), mutation (pairwise interchange of genes), and inversion (taking a random segment in a solution string and flipping it). They used a meta-GA to optimize crossover rate, inversion rate, and mutation rate. The individuals in the meta-GA population consisted of three integers in the range [0,20] (the overall search space was thus quite small, comprising 8000 combinations). An individual's fitness was defined as the quality of the best layout found when a GA was run with the parameters it embodied. The meta-GA was run on four test circuits, had a population size of 20, ran for 100 generations, and used uniform crossover (select parameter from either parent at random) and mutation (add random noise; no inversion used). They noted that crossover rate converged to values in the range 20-40\%, with little difference in best fitness as long as the rate was within this range. The mutation rate evolved by the meta-GA was 0.5-1.5\%, and the inversion rate was 0-30\%. The authors then adopted a crossover rate of 33\%, an inversion rate of 15\%, and a mutation rate of 0.5\%. These were used to run the optimizing GA on circuits of interest (different than those used in the meta-GA phase) and compare it with other techniques.

Another early work by \cite{fogel1991meta} described meta-evolutionary programming. Their flavor of evolutionary programming used mutation only to evolve solutions to two functions in $\mathbb{R}^2$. Mutation took a single parent and generated an offspring by adding  Gaussian random noise with zero mean and variance equal to $F(x,y)$---the function under investigation; this was applied to all members of the population. The meta-algorithm attached a perturbation term to each individual in the population, which was used as variance during mutation. This term was then evolved along with the solution. They compared both these algorithms (meta- and non-meta) with each other, and with a standard, Holland-style genetic algorithm, concluding that both versions of evolutionary programming outperformed the GA. They also concluded that the standard evolutionary-programming method attained better results over the two functions studied, but that the meta-version was more robust to changes in the function.

\cite{wu1995genetic} applied a genetic algorithm to nonlinear constrained mixed discrete-integer optimization problems, using a meta-GA to optimize population size, crossover probability, mutation probability, and crossover operator. The total number of parameter combinations was 19,200. The fitness of an individual in the meta-GA population was taken as the optimum found by a GA run with the parameters defined by the individual. Their findings showed insensitivity to crossover rate but high sensitivity to mutation rate. Four-point crossover outperformed one-, two-, and three-point crossover.  

\cite{hinterding1997adaptation} (see also \cite{eiben1999parameter}) noted that, ``it is natural to expect adaptation to be used not only for finding solutions to a problem, but also for tuning the algorithm to the particular problem.'' Their paper provided a short survey of adaptation techniques in evolutionary computation. They defined four categories of adaptation: static---constant throughout run and tuned externally; and dynamic, refined further into deterministic---parameter altered by a deterministic rule, adaptive---some feedback from the evolutionary algorithm determines the change in the parameter, and self-adaptive---the parameters to be adapted are encoded into the chromosomes and undergo crossover and mutation. They also defined four levels of adaptation: environment---such as changes to the fitness function (e.g., weights), population---parameters that apply to the entire population are adapted, individual---parameters held within an individual affecting only that individual, and component---parameters specific to a component or gene within an individual (such as self-adaptation of component-level mutation steps sizes and rotation angles in evolution strategies).  

\cite{ong2004meta} presented meta-Lamarckian learning in the context of memetic algorithms (MA), which incorporate local improvement procedures within traditional GAs. Their paper investigated the adaptive choice of local search (LS) methods to ensure robustness in MA search. In addition to Darwinian evolution they also studied Lamarckian learning, where the genotype reflected the result of improvement through placing the locally improved individual back into the population to compete for reproductive opportunities. They studied two adaptive meta-Lamarckian learning strategies, a heuristic approach based on subproblem decomposition, and a stochastic approach based on a biased roulette wheel. They tested their system on continuous parametric benchmark test problems and on a wing-design problem. They concluded that, ``the strategies presented are effective in producing search performances that are close to the best traditional MA with a LS chosen to suit the problem in hand. Given that such knowledge is often not available \textit{a priori}, this ability to tackle new problems in a robust way is of significant value.''

\cite{ramos2005logistic} proposed the utilization of logistic regression for tuning the parameters of a ``transgenetic'' algorithm---an evolutionary algorithm that deals, basically, with a population of chromosomes and a population of transgenetic vectors. They cited symbiogenesis as their inspiration, a theory of evolution according to which new cell organelles, new bodies, new organs, and new species arise from symbiosis, wherein independent organisms merge to form composites. The chromosomes of a transgenetic algorithm do not share genetic material directly. There are no crossover and mutation operations but rather transgenetic vectors that obtain and insert information into the chromosomes. They used logistic regression to set two main parameters of their algorithm (population size and maximum length of transgenetic vector that met certain constraints), their problem of interest being Traveling Salesman.  They showed that their algorithm outperformed a standard memetic algorithm.

\cite{brest2006self}, mentioned above, described an efficient technique for adapting control parameter settings associated with differential evolution (DE). DE uses a floating-point encoding for global optimization over continuous spaces, creating new candidate solutions by combining the parent individual and several other individuals of the same population. A candidate replaces the parent only if it has better fitness. DE has three parameters: amplification factor of the difference vector, crossover control parameter, and population size. In \cite{brest2006self}, the parameter control technique was based on the self-adaptation of the first two parameters, which were encoded within an individual's genome. Their testbed consisted of twenty-one benchmark functions from \cite{yao1999evolutionary}. They concluded that self-adaptive DE is better or comparable to the original DE and some other evolutionary algorithms they examined.

\cite{de2007parameter}---in his chapter in the book \textit{Parameter Setting in Evolutionary Algorithms} (\cite{lobo2007parameter})---provided a thirty-year perspective of parameter setting in evolutionary computation. He wrote that, ``It is not surprising, then, that from the beginning EA practitioners have wanted to know the answers to questions like:
\begin{quote}
\begin{itemize}
\item Are there optimal settings for the parameters of an EA in general?
\item Are there optimal settings for the parameters of an EA for a particular class of fitness landscapes?
\item Are there robust settings for the parameters of an EA that produce good performance over a broad range of fitness landscapes?
\item Is it desirable to dynamically change parameter values during an EA run?
\item How do changes in a parameter affect the performance of an EA?
\item How do landscape properties affect parameter value choices?''
\end{itemize}
\end{quote}

He went on to review static parameter-setting strategies, where he mentioned a two-level EA, the top level of which evolved the parameters of the lower-level EA. \cite{de2007parameter} stated that the, ``key insight from such studies is the robustness of EAs with respect to their parameter settings. Getting `in the ball park' is generally sufficient for good EA performance.'' Our study herein not only confirms this observation through numerous experiments, but also presents the novel finding that the ballpark can be quite large. Of dynamic parameter-setting strategies he opined that, ``it is difficult to say anything definitive and general about the performance improvements obtained through dynamic parameter setting strategies.'' He added, interestingly, ``My own view is that there is not much to be gained in dynamically adapting EA parameter settings when solving static optimization problems. The real payoff for dynamic parameter setting strategies is when the fitness landscapes are themselves dynamic ...''  \cite{de2007parameter}  also discussed the different aspects of setting various standard parameters: parent population size, offspring population size, selection, reproductive operators; adapting the representation; and parameterless EAs.  
 
\cite{kramer2010evolutionary} provided a survey of self-adaptive parameter control in evolutionary computation, where---as noted above---control parameters are added to the (evolving) genome. He complemented the taxonomy offered by \cite{eiben1999parameter}, dividing parameter setting into two main categories: tuning and control. Tuning was further divided into tuning by hand, tuning by design of experiments, and tuning by meta-evolution; while control was divided as previously into deterministic, adaptive, and self-adaptive. This paper mainly focused on function optimization techniques, such as the covariance matrix self-adaptation evolution strategy (CMSA-ES). A mention of meta-evolution noted that they, ``have to be chosen problem-dependent, which is the obvious drawback of the approach.'' He concluded with the observation that most theoretical work on self-adaptation concentrated on mutation, stating that, ``A necessary condition for the success of self-adaptation is a tight link between strategy parameters and fitness.''

\cite{eiben2011parameter} (see also \cite{smit2009comparing,smit2010parameter}) presented a conceptual framework for parameter tuning based on a three-tier hierarchy of: a problem, an evolutionary algorithm (EA), and a tuner. They argued that parameter tuning could be considered from two different perspectives, that of configuring an evolutionary algorithm by choosing parameter values that optimize its performance, and that of analyzing an evolutionary algorithm by studying how its performance depends on its parameter values. Furthermore, they distinguished between analyzing an evolutionary algorithm by studying how its performance depends on the problems it is solving, and analyzing an evolutionary algorithm by studying how its performance varies when executing independent repetitions of its run. They noted the existence of two types of parameters, \textit{qualitative} (e.g., crossover type) and \textit{quantitative} (e.g., crossover rate). They opined that, ``using tuning algorithms is highly rewarding. The efforts are moderate and the gains in performance can be very significant. Second, by using tuning algorithms one does not only obtain superior parameter values, but also much information about parameter values and algorithm performance. This information can be used to obtain a deeper understanding of the algorithm in question.'' The paper discussed a wide range of tuning algorithms, which they classified as sampling methods, model-based methods, screening methods, and meta-evolutionary algorithms. Of interest in their discussion of meta-evolutionary GAs was \cite{mercer1978adaptive}, an early (possibly first) though limited work; and the description of multi-objective meta-GAs, which tuned for more than a single objective, e.g., speed and accuracy. They opined that, ``parameter tuning in EC has been a largely ignored issue for a long time ... \textit{In the current EC practice parameter values are mostly selected by conventions, ad hoc choices, and very limited experimental comparisons.}''[italics added] This latter observation---with which we wholly concur---forms part of our motivation for the current study.

\cite{arcuri2013parameter} carried out ``the largest empirical analysis so far on parameter tuning in search-based software engineering.'' They performed experiments in the domain of test generation for object-oriented software using genetic algorithms. The objective was to derive sets of test cases (suites) for a given class, such that the test suite maximized a chosen coverage criterion while minimizing the number of tests and their length. A test case in this domain was a sequence of method calls that constructed objects and called methods on them. Because their goal was to study the effects of tuning, they analyzed all the possible combinations of the selected parameter values. They concluded that, ``tuning can improve performance, but default values coming from the literature can be already sufficient.''

\cite{bergstra2012random} studied neural networks, showing that random experiments were more efficient than grid experiments for hyper-parameter optimization in the case of several learning algorithms on several datasets. They wrote that, ``random experiments are more efficient because not all hyperparameters are equally important to tune ... Random experiments are also easier to carry out than grid experiments for practical reasons related to the statistical independence of every trial.'' This paper partly motivated our choice of random search in Section~\ref{sec-rascal}.

\cite{smit2010beating} is perhaps the most relevant paper to our current research, presenting a meta-EA called REVAC (Relevance Estimation and Value Calibration), which they used on a suite of 25 real-valued benchmark functions (real-parameter optimization functions defined for the CEC 2005 Special Session on Real-Parameter Optimization, including five unimodal functions and twenty multimodal functions; \cite{suganthan2005problem}). They chose to improve G-CMA-ES, which they considered a hard-to-improve evolutionary algorithm, cycling through \textit{parent selection-recombination-mutation-survivor selection-evaluation} over a population of G-CMA-ES parameter vectors. They were indeed successful in improving the algorithm's performance. 

Our aim is to go further, casting our net much wider in terms of problem domains, seeking to better understand parameter space.

\section{Software and Datasets}
\label{sec-GPs}

We chose to work with two very different evolutionary-algorithm packages: Distributed Evolutionary Algorithms in Python (DEAP) \cite{DEAP_JMLR2012}---which uses tree-based GP, and M4GP \cite{la_cava_william_genetic_2017}---which is a stack-based evolutionary algorithm. 
We ran our experiments on a cluster of 224 cores (Intel\textsuperscript{\textregistered} Xeon\textsuperscript{\textregistered} E5-2650L), with 2 threads per core.

DEAP, available at \texttt{github.com/DEAP}, comes with five sample problems:
\begin{enumerate}
\item Symbolic Regression, with data points generated from the quartic polynomial $x^4+x^3+x^2+x$. 
\item Even-Parity: find the parity, even or odd, of $n$ Boolean inputs (we set $n$ to 8).
\item Multiplexer 3-8: reproduce the behavior of an electronic multiplexer with 3 address bits (inputs) and 8 data lines (outputs).
\item Artificial Ant: evolve simple controllers---``artificial ants''---that are able to eat all the food located in a given two-dimensional, grid environment.
\item Spambase: return true if an email is spam, false otherwise. 
\end{enumerate} 
We performed a preliminary investigation of these five problems, essentially running DEAP with parameters tuned by hand (a common-enough undertaking in the EC community). We found that for the first four problems one can attain an accuracy level of close to 1, while for Spambase our preliminary investigation set the attainable accuracy level at 0.93. More detailed descriptions of each problem can be found on the DEAP website.

M4GP is entirely different, based on stack-based GP. This serves to reinforce our conclusions by running our experiments on two very different types of GP algorithms. M4GP uses a nearest centroid distance metric to make classifications, with each program producing multiple outputs.
We ran M4GP over problems from PMLB, a new publicly available dataset suite (accessibly hosted on GitHub) initialized with 165 real-world, simulated, and toy benchmark datasets for evaluating supervised classification methods \cite{Olson2017PMLB}. Note that PMLB focuses on classification benchmarks, whereas of the DEAP sample problems above only Spambase involves classification. Thus, our study includes different types of problems.

The preliminary investigation in this case consisted of delving into the data provided by the PMLB authors. ``Once the datasets were scaled,'' wrote \cite{Olson2017PMLB}, ``we performed a comprehensive grid search of each of the ML method's parameters using 10-fold cross-validation to find the best parameters (according to mean cross-validation balanced accuracy) for each ML method on each dataset. This process resulted in a total of over 5.5 million evaluations of the 13 ML methods over the 165 data sets.''

This available data saved us the need to run initial investigative experiments over PMLB. Each row of the 5.5-million table of results represents a single run, of a single ML algorithm, using a specific set of parameters; a row contains six columns: dataset name, classifier (machine learning algorithm), parameters, accuracy, macro-averaged F1 score, balanced accuracy. Given that one usually wants first and foremost to \emph{solve} a problem we focused on accuracy, specifically, balanced accuracy (the rightmost value of each row), which is a normalized version of accuracy that
accounts for class imbalance by calculating accuracy on a per-class basis, then averaging the
per-class accuracies \cite{Velez2007,urbanowicz2015exstracs}.

We composed two suites of ten datasets each: 1) 10 datasets for which a balanced accuracy of 1 was attained most frequently (Table~\ref{tab-pmlb-top}), and 2) 10 datasets whose average balanced accuracy was in the range $[0.9,0.95]$ (Table~\ref{tab-pmlb-inter}).

\begin{table}
\centering
\caption{PMLB results by \cite{Olson2017PMLB}. (a) ``Easier'' problems: 10 datasets for which a balanced accuracy of 1 was attained most frequently. (b)  ``Harder'' problems: 10 datasets whose average balanced accuracy was in the range $[0.9,0.95]$.
Shown for each problem: 
number of features,
number of classes,
and number of samples. 
}
\label{tab-pmlb}

\subtable[``Easier'' problems]{\label{tab-pmlb-top}{
\begin{tabular}{rccc}
problem     & features  & classes   & samples \\
mofn-3-7-10	& 10        & 2	        & 1324  \\
clean2	    & 168	    & 2         & 6598   \\
clean1	    & 168       & 2         & 476    \\
mushroom	& 22        & 2         & 8124   \\
irish	    & 5         & 2	        & 500    \\
agaricus-lepiota & 22   & 2	        & 8145   \\
corral	    & 6         & 2	        & 160     \\
xd6	        & 9	        & 2	        & 973    \\
mux6    	& 6         & 2	        & 128    \\
threeOf9	& 9	        & 2	        & 512     
\end{tabular}
}}
\hspace{10pt}
\subtable[``Harder'' problems]{\label{tab-pmlb-inter}{
\begin{tabular}{rccc}
problem     & features  & classes   & samples \\
breast-cancer-wisconsin & 30 & 2	& 569    \\
wdbc        & 30        & 2	        & 569    \\
tokyo1      & 44	    & 2	        & 959     \\
new-thyroid & 5	        & 3	        & 215     \\
spambase    & 57        & 2	        & 4601    \\
vote        & 16        & 2	        & 435    \\
soybean     & 35        & 18	    & 675    \\
house-votes-84 & 16	    & 2	        & 435     \\
breast-w    & 9         & 2	        & 699    \\
molecular-biology\_promoters & 58 &	2 & 106  
\end{tabular}
}}

\end{table}

\section{Searching for Parameters using a Meta-Genetic Algorithm}
\label{sec-megaton}
As done in a number of previous works discussed in Section~\ref{sec-prev}, we ran a meta-level genetic algorithm over the space of GP parameters, of which we identified five major ones: 
\begin{enumerate}
\item Population size ($\in \mathbb{N}$, $[100,3000]$),
\item Number of generations ($\in \mathbb{N}$, $[100,2000]$),
\item Crossover rate ($\in \mathbb{R}$, $[0,1]$),
\item Mutation rate ($\in \mathbb{R}$, $[0,1]$),
\item Tournament size ($\in \mathbb{N}$, $[3,100]$).
\end{enumerate}

\begin{figure}
\begin{center}
\includegraphics[width=0.9\textwidth]{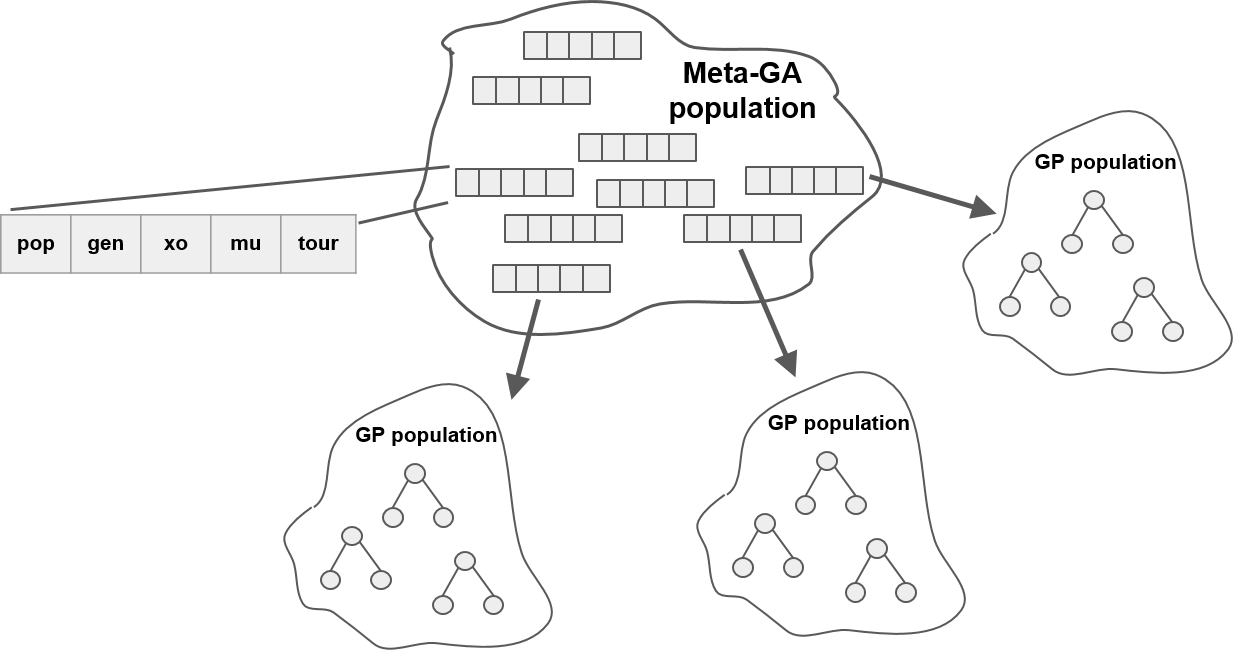}
\end{center}
\caption{The meta-genetic algorithm. The population consisted of simple linear genomes that contained five parameters. An individual's fitness was obtained by launching an entire GP evolutionary run with the parameters given in the genome. \textbf{pop}: population size,
\textbf{gen}: generation count,
\textbf{xo}: crossover probability,
\textbf{mu}: mutation probability,
\textbf{tour}: size of tournament for tournament selection.}
\label{fig-meta-ga}
\end{figure}

Figure~\ref{fig-meta-ga} shows a schematic of the meta-GA's workings. The meta-GA population comprised individuals with simple linear genomes that encoded the above five parameters as either integers or real values, respectively. An individual's fitness was obtained by launching entire GP evolutionary runs with the parameters given in the genome. The GP in question was either DEAP or M4GP. DEAP's goal was to solve the five problems listed in Section~\ref{sec-GPs} (regression, parity, mux, ant, spam), while M4GP was set lose on PMLB datasets. Each GP was run $n$ times, where $n$ was the number of problems it was set to solve (5 for DEAP, 10 for M4GP). Fitness was then a simple average of the $n$ highest fitness values obtained during the $n$ GP runs.

The meta-GA had a population size ranging from 100-300, ran for 50-160 generations, and used tournament selection with tournament size 3. Variance operators included: two-point crossover---with probability $p_{xo}=0.5$ perform crossover on a pair of parents by selecting two (of four possible) parameter boundaries and switching the parameters between them; and value mutation---with probability $p_{m}=0.2$ perform mutation on a single parent by selecting one of the five values at random, and mutating it by generating a random integer (population size, generation count, size of tournament) or real value (crossover probability, mutation probability) in the range specified above. 

We experimented with the meta-GA for approximately two months, performing \textit{tens of thousands of evolutionary runs}. We noted that numerous good parameter sets kept emerging, quite often appearing at random generation zero.
Hence, we designed several putative improvements to the algorithmic process. First, we tweaked parameters such as crossover and mutation, and also added elitism (2\%). It seemed that easier problems were causing the GP to move into local minima. To correct for this effect we introduced a weighted fitness function, where fitness was not computed as a simple average of the $n$ GP runs but rather a weighted one, with weights learned adaptively: examine every $m \in [1,5]$ generations the average fitness per problem attained by all GP runs over that particular problem; then increase weights of below-average problems and decrease weights of above-average problems.

Other tweaks are not described here for brevity. Suffice it to say that after numerous runs of the various versions of the meta-GA, we eventually concluded that there seemed to be numerous successful parameter sets.

\section{Searching for Parameters using Random Search}
\label{sec-rascal}
How rife is parameter space with good parameters, i.e., ones whose use by a GP run results in success (which needs to be carefully defined)? Given the findings by \cite{bergstra2012random} and the high cost grid search would incur in our case, we opted for random search. 

We generated parameter sets at random, ran the following sets of experiments, and recorded the successful ones:
\begin{enumerate}

\item DEAP over the 5 sample problems (regression, parity, mux, ant, spam). 
\begin{itemize}
\item Generate random parameter sets with parameters in the following ranges:
population size -- $[100,1000]$; 
generation count -- $[100,1000]$; 
crossover rate -- $[0,1]$; 
mutation rate -- $[0,1]$; 
tournament size -- $[3,30]$.

\item Total runs (i.e., random parameter sets generated, with 5 GP runs launched per each, attempting to solve all five problems): $2693$; number of successful parameter sets found: $110$.

\item Success criterion of a parameter set: accuracy of 0.97 attained for all 5 problems but Spambase, where the accuracy threshold was set to 0.93. 
\end{itemize}
Figure~\ref{fig-deap5-results} shows our results.

\begin{figure}
\centering
\subfigure[]{\label{fig-deap-top-pop}\includegraphics[width=0.45\textwidth]{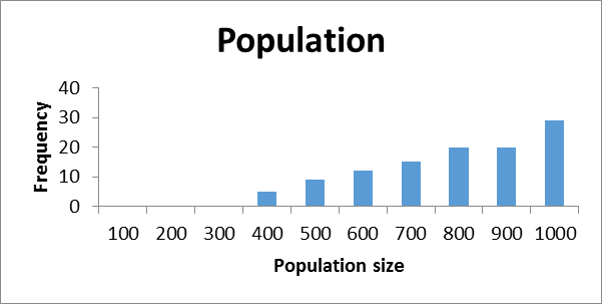}} \hfill
\subfigure[]{\label{fig-deap-top-gen}\includegraphics[width=0.45\textwidth]{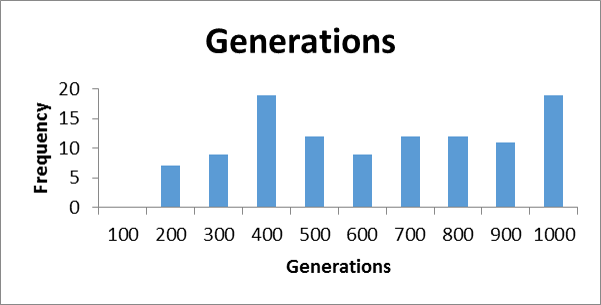}} 
\subfigure[]{\label{fig-deap-top-pop-gen}\includegraphics[width=0.45\textwidth]{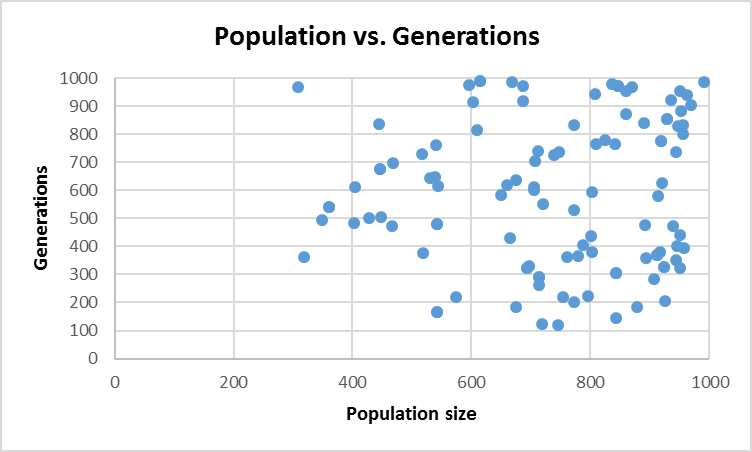}} \hfill
\subfigure[]{\label{fig-deap-top-xo}\includegraphics[width=0.45\textwidth]{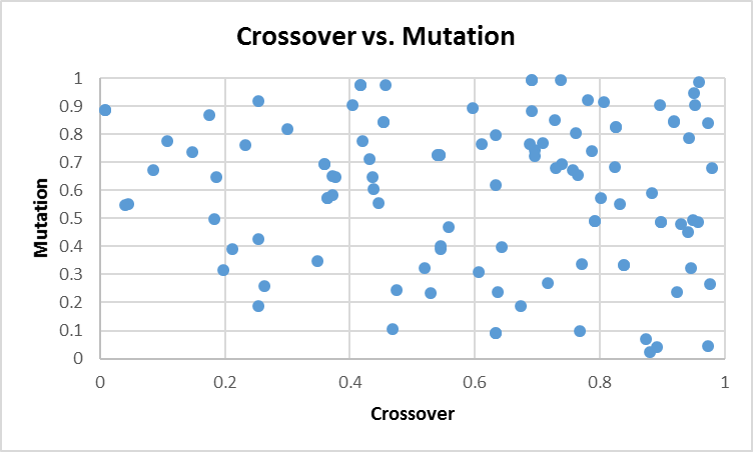}}
\subfigure[]{\label{fig-deap-top-tour}\includegraphics[width=0.45\textwidth]{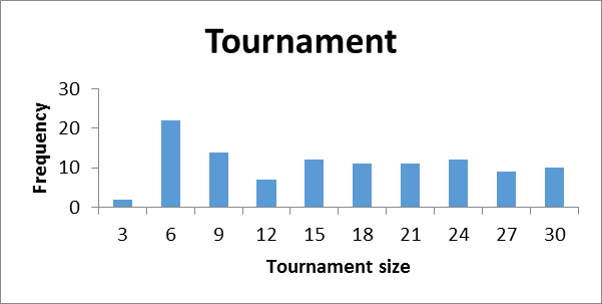}} 
\caption{DEAP run over 5 problems. 
         Shown are plots for the successful parameter sets found.}
\label{fig-deap5-results}
\end{figure}

\item M4GP over the 10 problems of Table~\ref{tab-pmlb-top}.
\begin{itemize}
\item Generate random parameter sets with parameters in the following ranges:
population size -- $[100,2000]$; 
generation count -- $[100,2000]$; 
crossover rate -- $[0,1]$; 
mutation rate -- $[0,1]$; 
tournament size -- $[3,30]$.

\item Total runs: $2610$; number of successful parameter sets found: $207$.

\item Success criterion of a parameter set: accuracy of 0.97 attained for all 10 problems. 
\end{itemize}
Figure~\ref{fig-pmlb-top-results} shows our results.

\begin{figure}
\centering
\subfigure[]{\label{fig-pmlb-top-pop}\includegraphics[width=0.45\textwidth]{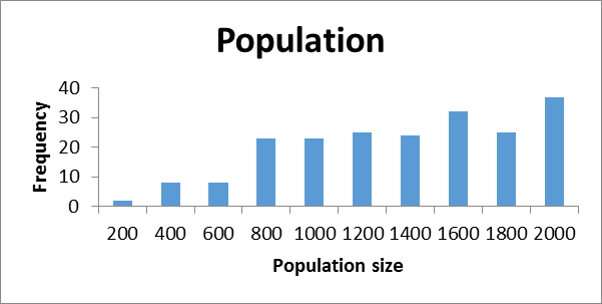}} \hfill
\subfigure[]{\label{fig-pmlb-top-gen}\includegraphics[width=0.45\textwidth]{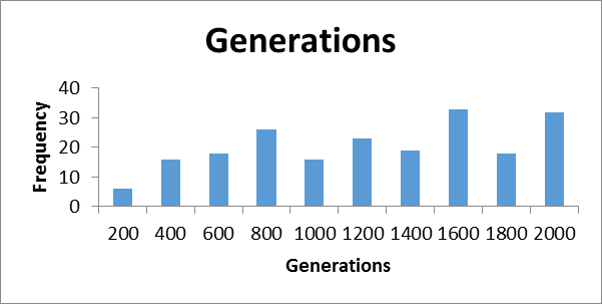}}
\subfigure[]{\label{fig-pmlb-top-pop-gen}\includegraphics[width=0.45\textwidth]{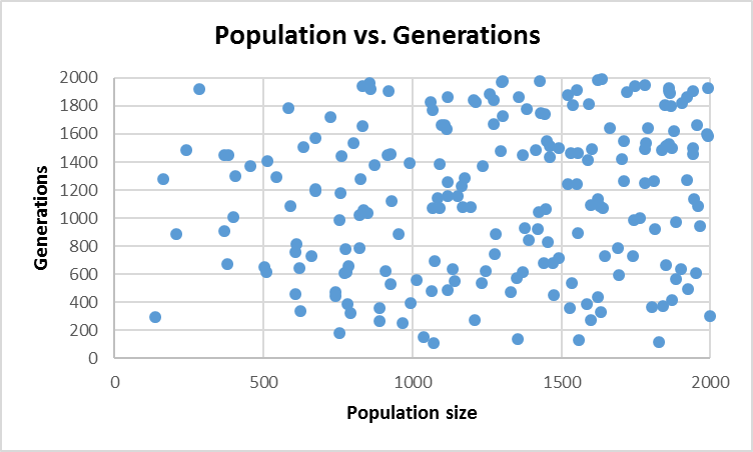}} \hfill
\subfigure[]{\label{fig-pmlb-top-xo}\includegraphics[width=0.45\textwidth]{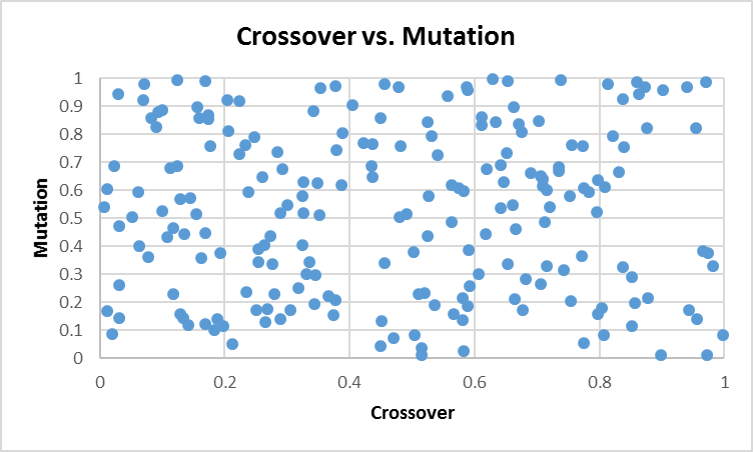}}
\subfigure[]{\label{fig-pmlb-top-tour}\includegraphics[width=0.45\textwidth]{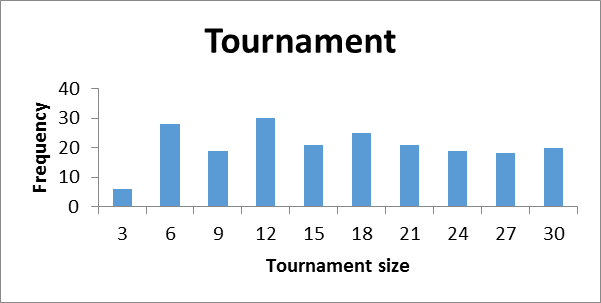}} 
\caption{M4GP over the 10 problems of Table~\ref{tab-pmlb-top}. 
         Shown are plots for the successful parameter sets found.}
\label{fig-pmlb-top-results}
\end{figure}

\item M4GP over the 10 problems of Table~\ref{tab-pmlb-inter}.
\begin{itemize}
\item Generate random parameter sets with parameters in the following ranges:
population size -- $[100,2000]$; 
generation count -- $[100,2000]$; 
crossover rate -- $[0,1]$; 
mutation rate -- $[0,1]$; 
tournament size -- $[3,30]$.

\item Total runs: $5432$; number of successful parameter sets found: $48$.

\item Success criterion of a parameter set: accuracy of 0.88 attained for all 10 problems. 
\end{itemize}
Figure~\ref{fig-pmlb-inter-results} shows our results.

\begin{figure}
\centering
\subfigure[]{\label{fig-pmlb-inter-pop}\includegraphics[width=0.45\textwidth]{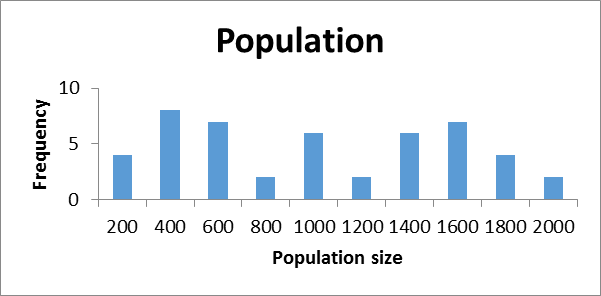}} \hfill
\subfigure[]{\label{fig-pmlb-inter-gen}\includegraphics[width=0.45\textwidth]{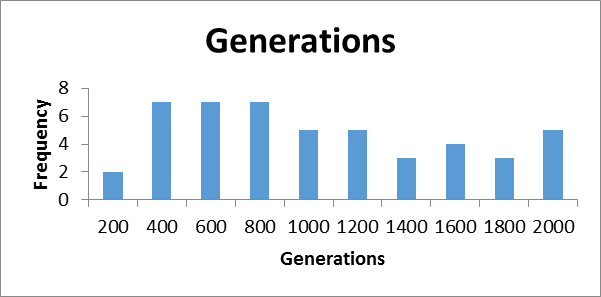}}
\subfigure[]{\label{fig-pmlb-inter-pop-gen}\includegraphics[width=0.45\textwidth]{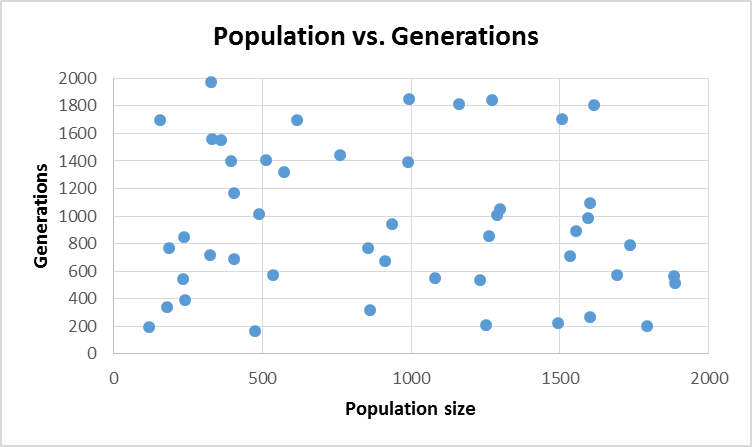}} \hfill
\subfigure[]{\label{fig-pmlb-inter-xo}\includegraphics[width=0.45\textwidth]{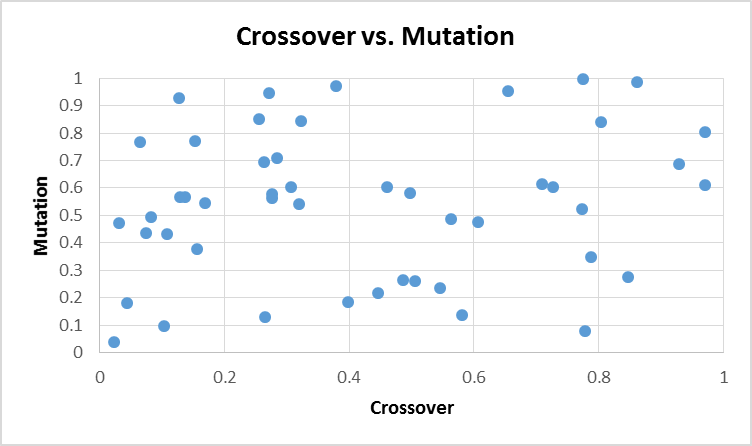}}
\subfigure[]{\label{fig-pmlb-inter-tour}\includegraphics[width=0.45\textwidth]{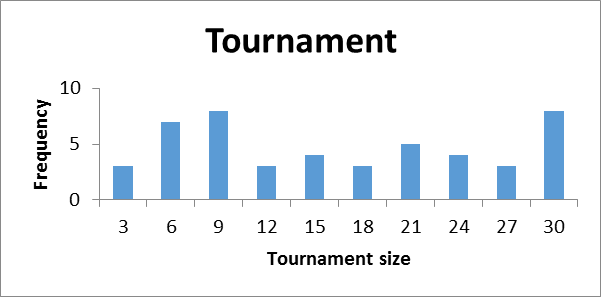}} 
\caption{M4GP over the 10 problems of Table~\ref{tab-pmlb-inter}. 
         Shown are plots for the successful parameter sets found.}
\label{fig-pmlb-inter-results}
\end{figure}

\end{enumerate}

The above experiments involved a total of \textit{93,615 GP runs}, each with a population size and generation count that could \textit{both be as high as 1000 or 2000}.

Perhaps evolution is not providing any added value and simple random search would suffice? This is a standard question one should ask, hence we performed the following:
\begin{enumerate}
\item DEAP:	For each of the $110$ ``good'' parameter sets found, generate $pop\_size \times gen\_count \times [1,5]$ random solutions and check how many of them pass the same 5-problem criterion employed above.

Result: none passed.

\textit{The total number of random solutions examined was 47,028,011.} 

\item M4GP:	For each of the $48$ ``good'' Table~\ref{tab-pmlb-inter} parameter sets found, generate $pop\_size \times gen\_count \times [1,10]$ random solutions and check how many of them pass the same 10-problem criterion employed above.\footnote{Note: The number of actual solutions generated per test point could be less than 5 or 10, respectively, if the criterion failed early on.}  

Result: none passed.

\textit{The total number of random solutions examined was 205,377,967.} 
\end{enumerate}
Random solutions were generated by using the respective software package to generate generation zero of size $pop\_size \times gen\_count$. This not only saved programming time but also prevented any bias vis-a-vis the GP experiments. 

\section{Concluding Remarks}
\label{sec-conc}
We performed what is arguably one of the most extensive EC experiments.
Studying our results in Figures~\ref{fig-deap5-results},~\ref{fig-pmlb-top-results}, and~\ref{fig-pmlb-inter-results}, a clear trend emerges: There is very little trend. We clearly see that good parameters range over the entire spectrum, somewhat in contrast with common lore. At least for 25 the problems studied herein we note that:
\begin{itemize}
\item Population size need not be maximal.
\item Generation count need not be maximal.
\item At most, population size and generation count should not both be very low. We remark that recent evolutionary findings suggest the use of fewer generations \cite{EvolutionFaster}.
\item Tournament size need not be in the commonly used range of $3\mbox{-}7$.
\item Crossover rate can be high, low, or intermediate.
\item Mutation rate can be high, low, or intermediate.
\item Crossover-mutation pairs show no tendency to aggregate anywhere.
      Crossover-mutation pairs should perhaps not both be low (which makes sense given that an evolutionary algorithm requires inter-generational variation).
\end{itemize}

\begin{figure*}
\centering
\subfigure[]{\label{fig-symreg}\includegraphics[width=0.32\textwidth]{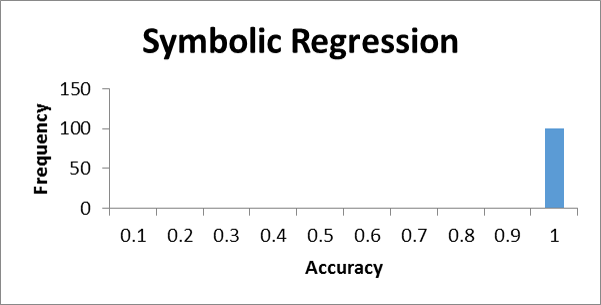}} \hfill
\subfigure[]{\label{fig-evenparity}\includegraphics[width=0.32\textwidth]{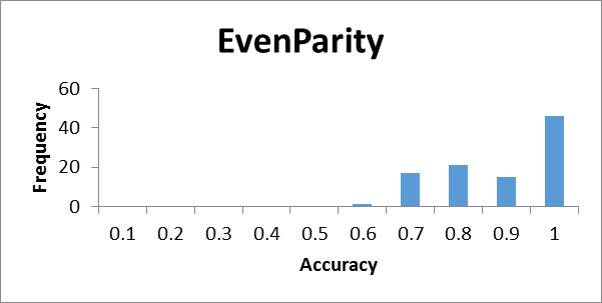}} \hfill
\subfigure[]{\label{fig-mux}\includegraphics[width=0.32\textwidth]{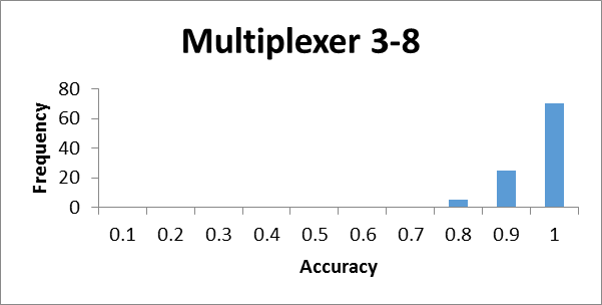}} 

\subfigure[]{\label{fig-ant}\includegraphics[width=0.32\textwidth]{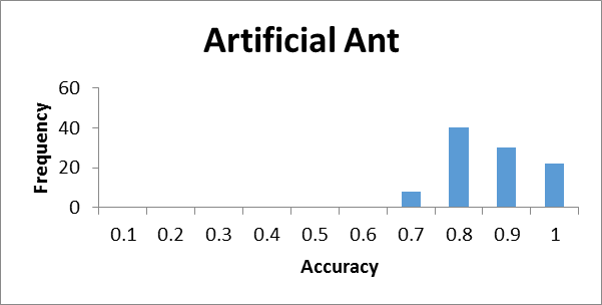}} \hfill
\subfigure[]{\label{fig-spambase}\includegraphics[width=0.32\textwidth]{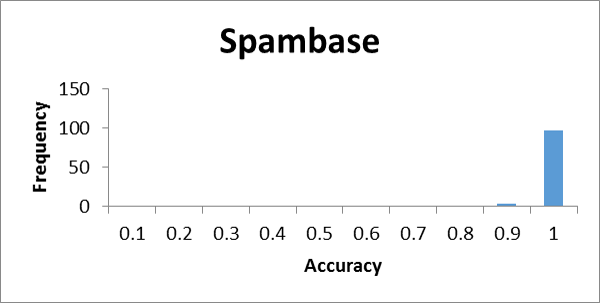}} \hfill
\subfigure[]{\label{fig-breast}\includegraphics[width=0.32\textwidth]{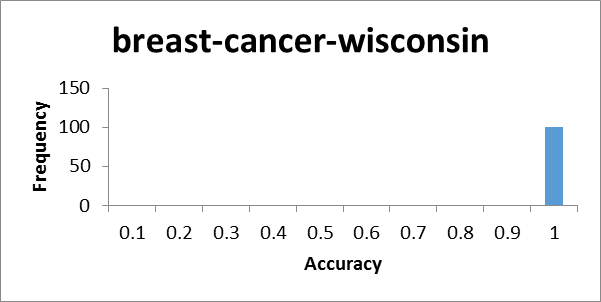}} 

\subfigure[]{\label{fig-wdbc}\includegraphics[width=0.32\textwidth]{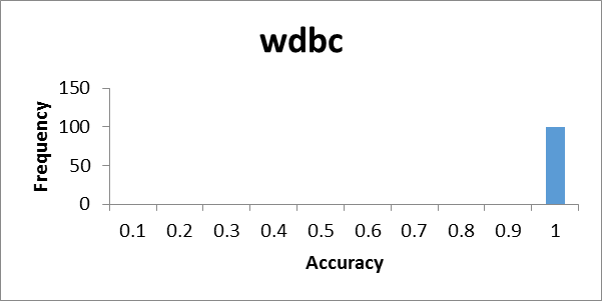}} \hfill
\subfigure[]{\label{fig-tokyo1}\includegraphics[width=0.32\textwidth]{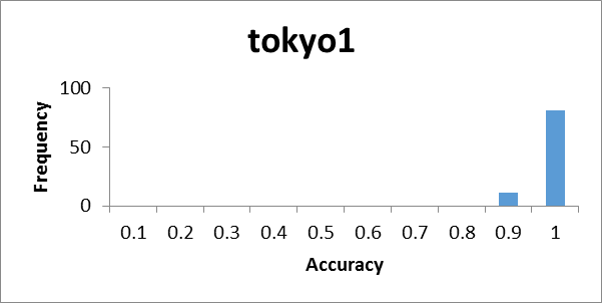}} \hfill
\subfigure[]{\label{fig-thyroid}\includegraphics[width=0.32\textwidth]{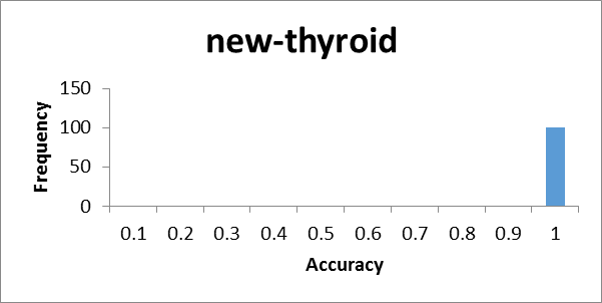}} 

\subfigure[]{\label{fig-mushroom}\includegraphics[width=0.32\textwidth]{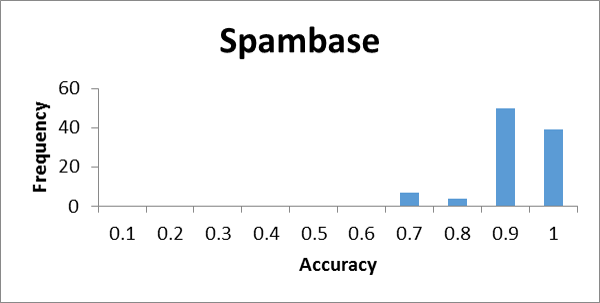}} \hfill
\subfigure[]{\label{fig-vote}\includegraphics[width=0.32\textwidth]{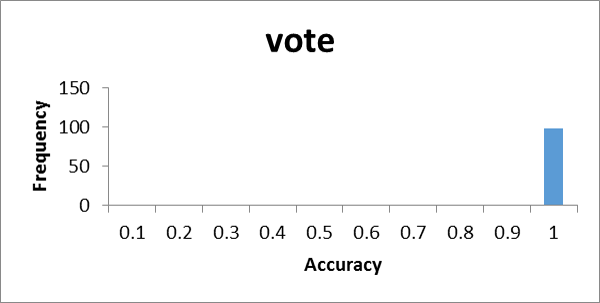}} \hfill
\subfigure[]{\label{fig-soybean}\includegraphics[width=0.32\textwidth]{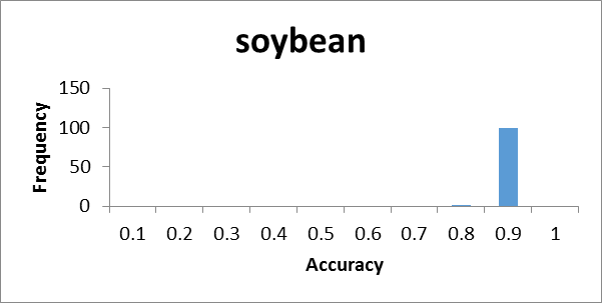}} 

\subfigure[]{\label{fig-house}\includegraphics[width=0.32\textwidth]{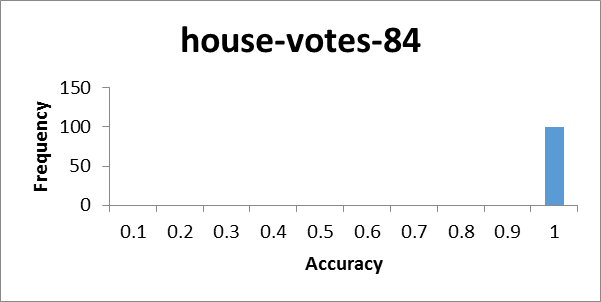}} \hfill
\subfigure[]{\label{fig-breast-w}\includegraphics[width=0.32\textwidth]{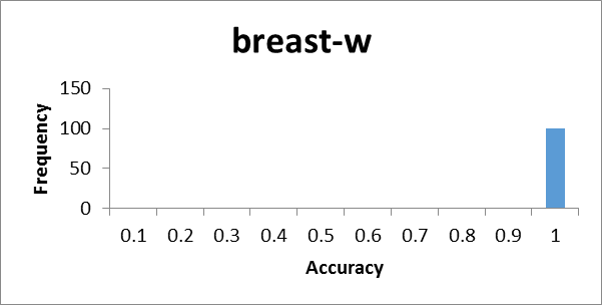}} \hfill
\subfigure[]{\label{fig-molecular}\includegraphics[width=0.32\textwidth]{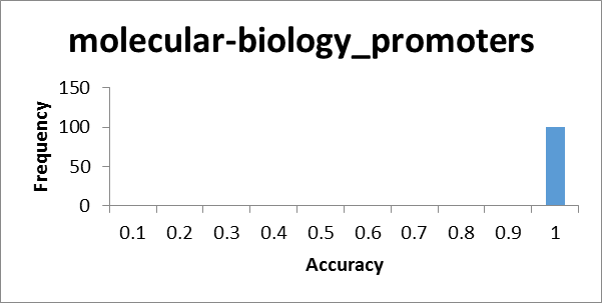}} 
\hspace*{\fill} \hspace*{\fill} \hspace*{\fill}  \hspace*{\fill}  \hspace*{\fill}  \hspace*{\fill}  \hspace*{\fill}  \hspace*{\fill} \hspace*{\fill} \hspace*{\fill} \hspace*{\fill} \hspace*{\fill} \hspace*{\fill} \hspace*{\fill} \hspace*{\fill} \hspace*{\fill} \hspace*{\fill} 

\caption{Results for single-problem test. For each of the above 15 problems, 100 random parameter sets were generated, and a GP run executed for every one, recording the best accuracy attained. (a)-(e): DEAP,  (f)-(o):M4GP}
\label{fig-single-problem-tests}
\end{figure*}

The yield (percentage) of good random parameter sets may seem relatively small, but that is not the point, rather, \textit{our major observation concerns the diversity in parameter space}. Moreover, this yield
will likely be far higher in most situations encountered by evolutionary-algorithm practitioners, as we were aiming at a broad spectrum of problems---quite a high and somewhat unconventional bar---rather than at the more common case of a single problem. To test this latter hypothesis we ran two additional sets of experiments: 
\begin{enumerate}
\item DEAP:	For each of the 5 problems, generate 100 random parameter sets and perform a GP run for each, recording the best accuracy attained.
\item M4GP:	For each of the 10 problems of Table~\ref{tab-pmlb-inter}, generate 100 random parameter sets and perform a GP run for each, recording the best accuracy attained.
\end{enumerate}
Figure~\ref{fig-single-problem-tests} shows the results of these $1500$ runs.

Our findings suggest that perhaps one need not always spend too much time and resources on tuning hyper-parameters and that random search is a good choice for such tuning. Moreover, robustness to hyper-parameter tuning is a desired quality of an evolutionary algorithm and if one's algorithm requires very specific parameters, the chances of finding them are slim; this would essentially be a needle-in-a-haystack situation in hyper-parameter space. 


\section*{Acknowledgments}
We thank William La Cava, Randal Olson, and Ryan Urbanowicz for helpful discussions.
This research was supported by National Institutes of Health grant no. AI116794. 

\bibliography{meta-ga}{}
\bibliographystyle{spmpsci}

\end{document}